\documentclass[conference]{IEEEtran}
\usepackage{cite}
\usepackage{amsmath,amssymb,amsfonts}
\usepackage{algorithmic}
\usepackage{graphicx}
\usepackage{textcomp}
\usepackage{hyperref}
\usepackage{makecell}
\usepackage{caption}
\usepackage{float}

\def\BibTeX{{\rm B\kern-.05em{\sc i\kern-.025em b}\kern-.08em
    T\kern-.1667em\lower.7ex\hbox{E}\kern-.125emX}}

\begin{document}

\title{Does CLIP perceive art the same way we do?}

\author{\IEEEauthorblockN{Andrea Asperti}
\IEEEauthorblockA{\textit{Dept. of Informatics (DISI)} \\
\textit{University of Bologna}\\
Bologna, Italy \\
andrea.asperti@unibo.it}
\and
\IEEEauthorblockN{Leonardo Dessì}
\IEEEauthorblockA{\textit{Dept. of Informatics (DISI)} \\
\textit{University of Bologna}\\
Bologna, Italy \\
leonardo.dessi@studio.unibo.it}
\and
\IEEEauthorblockN{Maria Chiara Tonetti}
\IEEEauthorblockA{\textit{Dept. of Informatics (DISI)} \\
\textit{University of Bologna}\\
Bologna, Italy \\
mariachiara.tonetti2@unbo.it}
\and
\IEEEauthorblockN{Nico Wu}
\IEEEauthorblockA{\textit{Dept. of Informatics (DISI)} \\
\textit{University of Bologna}\\
Bologna, Italy \\
nico.wu@studio.unibo.it}
}

\maketitle

\begin{abstract}
Multimodal systems and Large Language Models have shown remarkable capabilities in text-based reasoning, yet their capacity to perceive and interpret visual art remains uncertain. This study examines how CLIP “sees” and understands artworks by comparing their responses to human- and AI-generated paintings in the European tradition from the Renaissance onward. The analysis focuses on its ability to identify style, period and cultural context, as well as potential biases in its perception, evaluated against human judgments.
\end{abstract}


\begin{figure*}[h]
  \includegraphics[width=\textwidth]{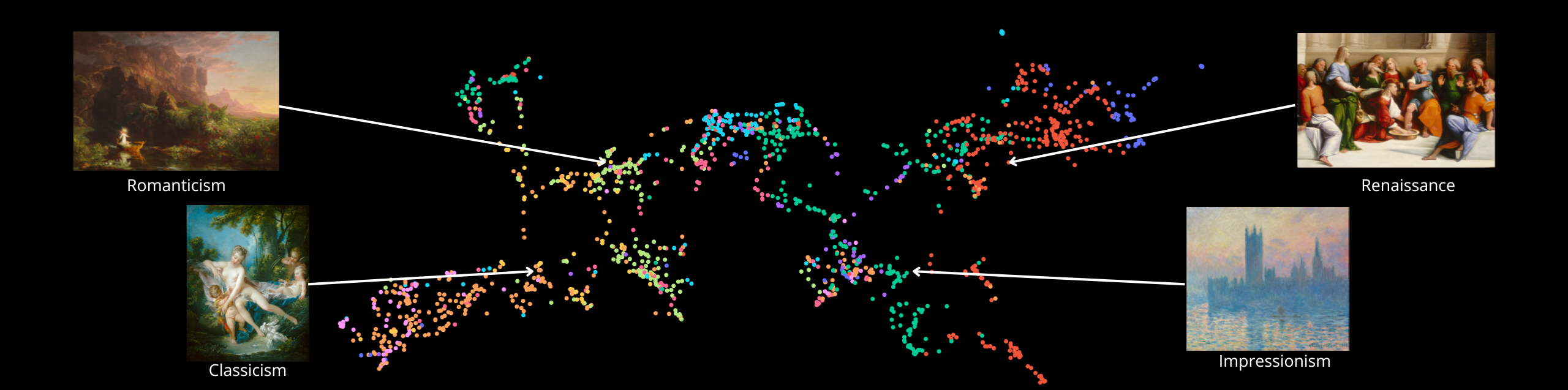}
  \caption{3D UMAP projection of image embeddings extracted by the CLIP ViT-L/14 model from the National Gallery of Art.}
  \label{fig:teaser}
\end{figure*}
\section{Introduction}
In recent years, multimodal models have reshaped the landscape of machine perception and understanding, with CLIP (Contrastive Language–Image Pretraining) \cite{CLIP} standing out as one of the most influential. Trained on hundreds of millions of image–text pairs, CLIP has demonstrated remarkable capabilities across a broad range of tasks, including zero-shot classification \cite{CLIPMulti,CLIPSAM,CLIP-zero-shot}, image retrieval \cite{CAMIR,Retrieval_CLIP,CLIP_ECAI2024}, re-identification \cite{CLIP-ReID,CLIP_GAP}, prompt-based generation \cite{CLIP_prior_guidance,Distilling_CLIP}, and semantic search \cite{DBLP:conf/colins/LytvynPRKG24,SgVA-CLIP,CLIP-MUSED}. Its success has made it a foundational component in many state-of-the-art generative models and vision-language pipelines. In particular, CLIP’s ability to align visual and textual modalities has been widely adopted as a guiding mechanism in tasks ranging from text-to-image synthesis (e.g., GLIDE \cite{GLIDE}, DALL·E \cite{DALLE2}, Stable Diffusion \cite{stable-diffusion,stable_diffusion_3_5_large_t2i}) to creative applications such as style transfer and visual storytelling.

Yet, despite its ubiquity, the nature and limits of CLIP’s perceptual alignment remain underexplored. While CLIP is optimized to match images with descriptive captions, it is less clear whether this alignment genuinely reflects human perception — especially in complex, subjective, or culturally embedded domains like art. Its capacity to identify what is depicted in an image is remarkable, but its understanding of how that content is rendered — including factors such as artistic style, visual coherence, or historical context — is far less understood. 

In this paper, we investigate CLIP’s perceptual capabilities in the domain of visual art, focusing on both human-made and AI-generated paintings. We treat CLIP’s vision encoder as a fixed perceptual system, without any fine-tuning or modification, in analogy to the human sensory apparatus.
Our goal is to assess how well CLIP captures not only semantic content but also stylistic attributes, temporal signals, and visual deformations. Through a series of probing tasks and analyses, we evaluate CLIP’s representations along multiple interpretive axes, including scene type, artistic style, historical period, and the presence of visual artifacts. To this end, we leverage two richly annotated datasets: classical works from the \href{https://www.nga.gov/open-access-images.html}{National Gallery of Art} of Washington \cite{nga_real_dataset} and synthetic paintings from the AI-Pastiche \cite{AI-pastiche} collection.

This study serves as a stepping stone toward a broader question: can large vision-language models, such as CLIP, form something akin to an aesthetic sense? Do they internalize representations of style, harmony, or beauty, and if so, are these grounded in visual abstraction, statistical regularities, or biases in their training data?

We offer both a conceptual and empirical investigation into these questions. Our findings reveal a consistent gap between CLIP’s textual associations and its ability to perceive visual nuance. While the model performs well on broad semantic alignment, it struggles with stylistic subtleties, the attribution of artistic periods, and the detection of visual defects in generative imagery. These limitations point to a deeper issue: despite its multimodal power, CLIP lacks a robust internal model of aesthetic form.

By combining metadata-driven evaluation with perceptual baselines, our work presents a structured critique of CLIP’s performance in aesthetic domains. In doing so, we highlight the need for more interpretable and perceptually grounded multimodal systems, particularly as models like CLIP are increasingly used to guide, evaluate, and curate creative content. 

\section{Related Works}
\label{sec:related}
Several recent studies have examined the potential and limitations of CLIP’s vision mechanism, particularly how well its image embeddings can be retrieved and exploited via textual prompts. Concerns about CLIP’s ability to fully grasp prompt semantics were raised in \cite{CLIP_semantic_knowledge}. In \cite{CLIP_limitations}, controlled experiments in multi-object contexts revealed notable biases: the image encoder favors larger objects, while the text encoder prioritizes objects mentioned first. CLIP’s robustness has also been investigated. \cite{CLIP_adversarial} showed that CLIP-based methods for detecting AI-generated images are vulnerable to white-box attacks, while \cite{CLIP_robustness} offered a broader evaluation, finding that despite limitations, CLIP models are generally more robust to visual factor variations than traditional ImageNet models.


Another line of research has investigated the use of CLIP as a guidance technique for image generation. Findings in \cite{CLIP_closer_look} indicate that while CLIP embeddings contribute to aesthetic quality, they have limited influence on maintaining consistency between subject and background. Comparable limitations are discussed in  \cite{Semantic_Diffusion_Guidance}, which emphasizes the need for a more unified framework integrating text- and image-guided synthesis. However, none of these studies considers guidance for style replication, and the results presented here suggest that CLIP offers limited benefit for this purpose and may, in some cases, be counterproductive.

 The possibility of understanding CLIP internal representations by inverting them with traditional gradient ascent techniques \cite{Vevaldi15,inverting_CNN,dreaming_to_distill} has been explored in \cite{inverting_CLIP}. 
 A gradient ascent technique, not based on pixel optimization but on a set of RGBA Bezier curves, is investigated in \cite{CLIPDraw}.
 All these studies seem to highlight the fact that only basic semantic information is encoded in CLIP's embeddings.
 
 In light of the previous limitations, several works have been devoted to improving CLIP's performance, enhancing its downstream generalization ability, and reducing the modality gap between text and images. 
 Adapters technique allows for a lightweight fine-tuning of the model through the insertion of suitable modules.
 Examples along this direction are CLIP-adapter 
 \cite{CLIP-adapter},
 TIP-adapter \cite{TIP-adapter}, LIxP \cite{context_aware_pretraining} or APE \cite{APE}.
 Given that our aim is to explore CLIP’s native perceptual abilities in the context of art, we consider adaptation techniques to be somewhat misaligned with the spirit of our investigation.
 A line of investigation closer to our objectives consists of addressing potential perception issues by focusing on subspaces through suitable projections, as outlined in \cite{CLIP_subspace}. However, our preliminary investigations in this direction did not yield interesting
 results. 

 A different research direction consists of improving the discrimination and retrieval capacities of multimodal systems through prompt engineering. A common followed approach, e.g. in CoOp \cite{CoOp} or CoCoOp \cite{CoCoOp}, consists of integrating the tokenization of the prompt with a set of learnable vectors learned by gradient descent. We tested this technique in the case of style classification, without noticeable improvements.
 


\section{Methodology and Data}
\label{sec:methodology}
Our aim is to investigate CLIP's ability to extract high-level semantic and stylistic information from paintings and to evaluate its perceptual capabilities across multiple dimensions, including content, scene understanding, artistic style, and the presence of visual deformations or artifacts.

We conducted our analyses on two richly annotated datasets: a subset of the freely available National Gallery of Art collection in Washington, and the AI-Pastiche dataset \cite{AI-pastiche}—a collection of 953 AI-generated paintings produced by 12 different models from 73 carefully crafted prompts, covering a wide range of major artistic styles. 
The motivation for comparing human and AI-generated artworks lies in our goal to assess whether CLIP's vision system can perceive the nuanced differences between them, including the limitations of AI models in accurately replicating artistic styles.

For the paintings from the National Gallery of Art, we leverage the available metadata to evaluate CLIP's ability to associate each work with a descriptive summary and to understand its artistic style. We employ the Uniform Manifold Approximation and Projection (UMAP) algorithm \cite{UMAP} as a visualization tool for exploring the shared image-text latent space of CLIP. An example of the resulting visualization, where the embeddings of the paintings are color-coded by their respective styles, is shown in Figure~\ref{fig:teaser}.

A similar investigation is conducted on the AI-Pastiche dataset, comparing the generated images to both their textual prompts and to the intended artistic styles. This analysis is inherently more complex, as any mismatch between image and prompt could result from either of the two models involved: the image generator, which may fail to accurately follow the prompt, or CLIP, which may fail to correctly identify the intended content or style. To unravel these factors, we draw on a set of user surveys conducted by the creators of the AI-Pastiche dataset. These surveys assess the perceived "authenticity" of AI-generated artworks, their adherence to the prompt, and the presence of visible artifacts or deformations.

Specifically, we use CLIP to evaluate how closely each generated image aligns with its corresponding prompt and compare these results with human evaluations. In addition, we test CLIP’s capacity to identify visual defects, such as distortions, inconsistencies, or artifacts commonly produced by generative models.

Most of our investigations were conducted using multiple versions of CLIP, with the secondary goal of comparing their relative performance.

\section{Experiments on human artworks}

The experiments in this section were conducted on a subset of the National Gallery of Art in Washington.
We designed two distinct experiments. The first aims to assess CLIP's ability to associate each image with a summary of its corresponding description. The second focuses on evaluating CLIP's capacity to distinguish between different artistic styles.

\subsection{Image-description alignment}
\label{sec:description similarity}
In the first experiment, a limitation arose from CLIP’s 77-token input cap, as painting descriptions from the National Gallery of Art often exceeded this length. To address this, concise summaries (x) were generated using ChatGPT-4o-mini, ensuring that content, style, and period information from the original description was preserved. The actual image-generation prompt is provided in the extended version \cite{does_clip_arxiv}. For each image–summary pair ⟨x, s⟩, we computed the cosine similarity between  $CLIP_{image}(x)$ and $CLIP_{text}(s)$ and used this value as a ranking score to compute recall at different thresholds. 
The experiments were repeated for several different versions of CLIP available in the OpenAI library\cite{CLIP}; results are
reported in Table \ref{tb:NGD_result_table}.
\begin{center}
\begin{table}[htbp]\small
\begin{tabular}{l|ccc}
\textbf{Model}          &\textbf{ recall@1} & \textbf{recall@5} & \textbf{recall@10} \\ \hline
RN50           & 0.663 & 0.915 & 0.966 \\
RN101          & 0.693 & 0.926 & 0.966 \\
RN50x4         & 0.741 & 0.946 & 0.978 \\
RN50x16        & 0.791 & 0.964 & 0.988 \\
RN50x64        & \textbf{0.828} & 0.97  & 0.99  \\
ViT-B/32       & 0.678 & 0.925 & 0.97  \\
ViT-B/16       & 0.709 & 0.928 & 0.969 \\
ViT-L/14       & 0.794 & 0.972 & 0.989 \\
ViT-L/14@336px & 0.814 & \textbf{0.974} & \textbf{0.991} 
\end{tabular}
\caption{Summary-image alignment for NGAD images.}
\label{tb:NGD_result_table}
\end{table}
\end{center}

The results show a consistent improvement in performance as the capacity of the CLIP models increases. 
Overall, while the ResNet-based RN50x64 yields the best recall@1 scores in this experiment, ViT-L/14@336px performs comparably, especially at higher recall thresholds. This suggests that both architectural complexity and input resolution play a crucial role in enhancing the image-text alignment capabilities of CLIP, particularly in tasks involving fine-grained associations, such as matching painting summaries with artworks.




\subsection{Style Recognition}
\label{sec:style_NGD}

In this second experiment, we assess CLIP’s ability to associate artworks with their corresponding artistic styles. For each unique style present in the dataset, we generated a fixed textual prompt in the form “an artwork in [style] style”. Using CLIP’s image and text encoders, we computed normalized embeddings for both modalities. The cosine similarity between text and image embeddings was then calculated to produce a similarity matrix, capturing the degree of alignment between each image and every style prompt, at different recall thresholds.


Table \ref{tb:style_result_table_ngd} reports the results obtained across a range of CLIP architectures. Among all tested models, \texttt{ViT-L/14@336px} achieved the best performance, with a recall@1 of 0.457 and a recall@5 of 0.831. However, the results across all models remain moderate, highlighting the increased complexity of style recognition compared to textual description alignment.

\begin{table}[htbp]\small
\begin{tabular}{l|cccc}
model & recall@1 & recall@2 & recall@3 & recall@5 \\ \hline
RN50 & 0.354 & 0.546 & 0.662 & 0.801 \\
RN101 & 0.379 & 0.577 & 0.674 & 0.78 \\
RN50x4 & 0.344 & 0.504 & 0.611 & 0.772 \\
RN50x16 & 0.373 & 0.581 & 0.679 & 0.786 \\
RN50x64 & 0.343 & 0.516 & 0.627 & 0.766 \\
ViT-B/32 & 0.316 & 0.467 & 0.585 & 0.737 \\
ViT-B/16 & 0.349 & 0.506 & 0.622 & 0.766 \\
ViT-L/14 & 0.4 & 0.577 & 0.697 & 0.795 \\
ViT-L/14@336px & \textbf{0.457} & \textbf{0.632} & \textbf{0.716} & \textbf{0.831}
\end{tabular}
\caption{Style recognition scores on NGAD.}
\label{tb:style_result_table_ngd}
\end{table}

\begin{figure}[htbp] 
    \centering 
    \includegraphics[width=1\linewidth]{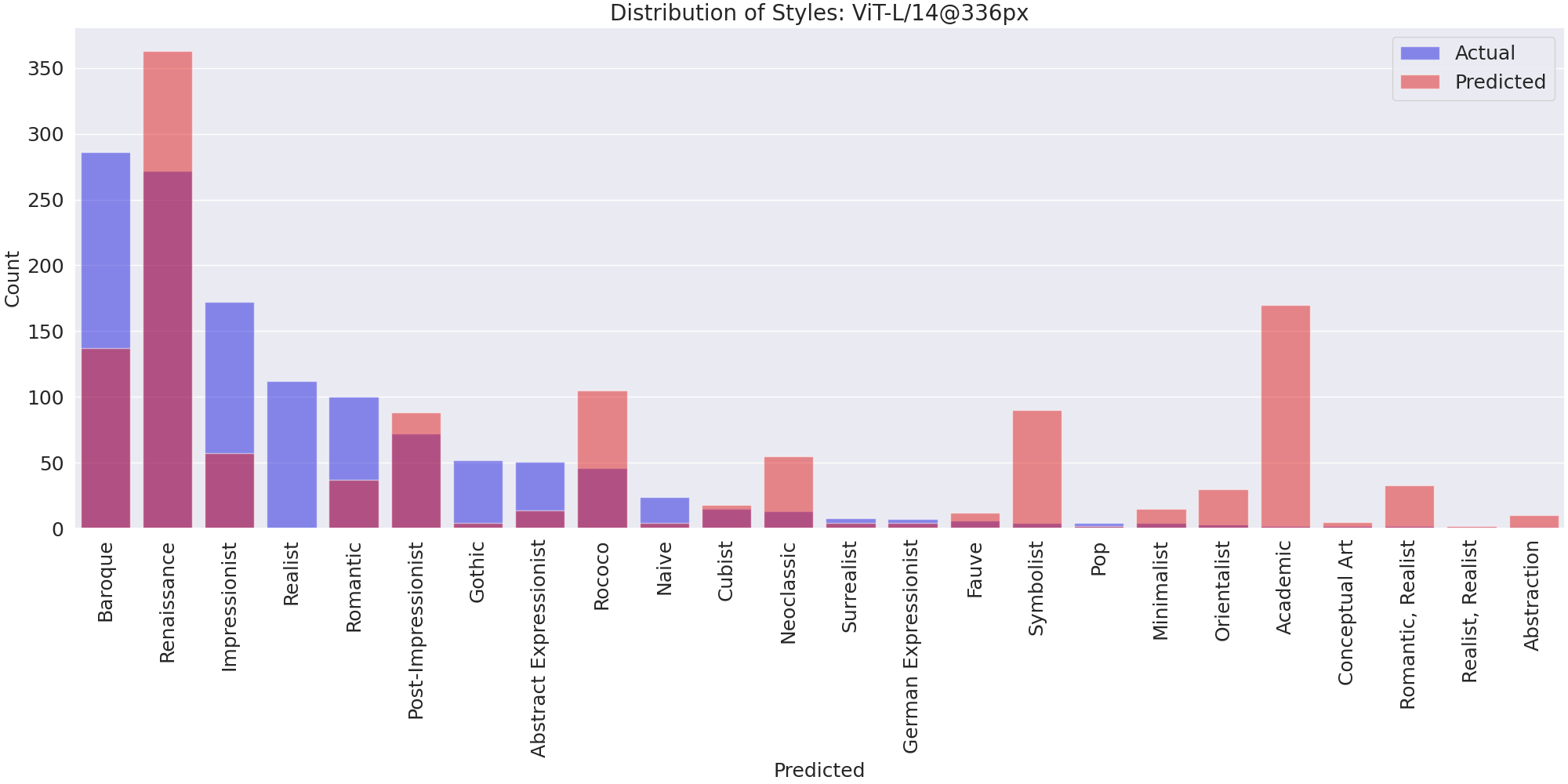}
    \caption{Comparison between actual and predicted distribution of styles in NGAD.}
    \label{fig:histogram_style_NGD} 
\end{figure}

In Figure \ref{fig:histogram_style_NGD}, we compare the distribution of true and predicted styles across the dataset. This analysis highlights notable disparities, reflecting both the imbalance present in the dataset and the most frequent mistakes in the model's predictions.
These findings suggest that while CLIP demonstrates a degree of sensitivity to artistic style, its current representations are not fully suited for tasks requiring a nuanced understanding of artistic conventions or visual grammar, warranting further refinement or supervision for such objectives.

\begin{figure}[htbp] 
    \centering 
    \begin{tabular}{ccc}\includegraphics[height=.33\columnwidth]{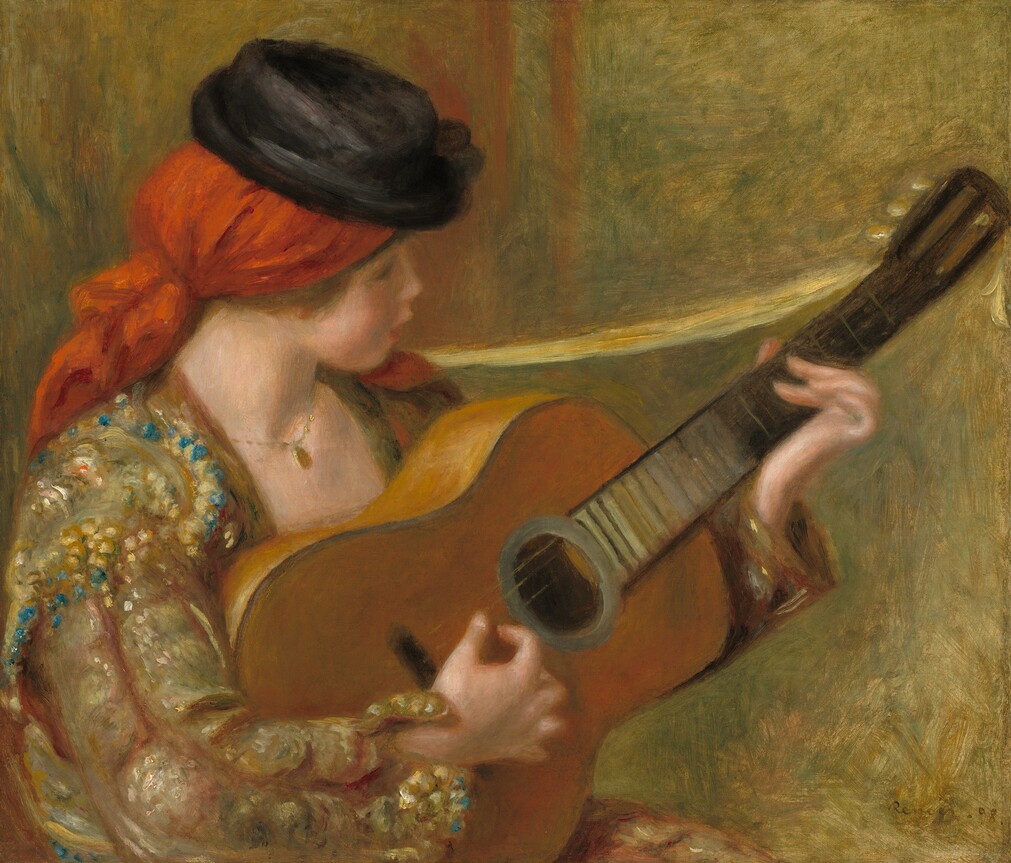} & \includegraphics[height=.33\columnwidth]{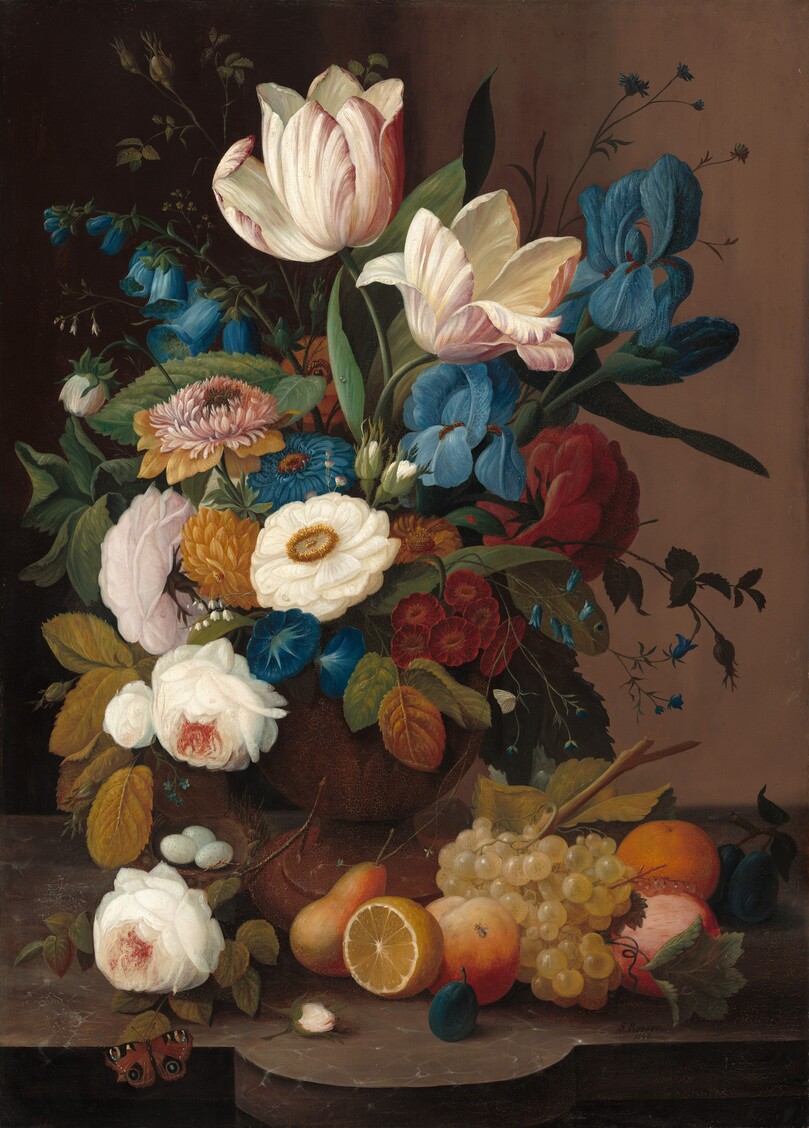} & \includegraphics[height=.33\columnwidth]{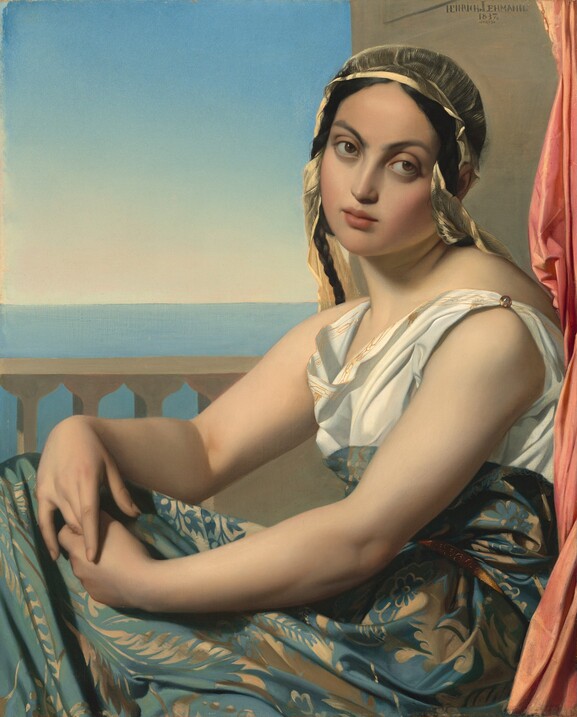} \\
            \makecell{True: Impressionist\\Pred: Renaissance} &
    \makecell{True: Realist\\Pred: Rococo} &
    \makecell{True: Orientalist\\Pred: Neoclassic}
    \end{tabular} 
    \caption{Examples of misclassifications in style recognition} 
    \label{fig:style_misclassifications} 
\end{figure}

A qualitative inspection of misclassified examples offers further insight into CLIP’s limitations in style recognition. Figure~\ref{fig:style_misclassifications} showcases three representative failure cases. Each image is shown with its true style label and the incorrect prediction made by the best-performing model.
These examples illustrate how overlapping visual features or contextual ambiguity can lead to misclassifications. 

\subsection{Semantic Relationships in the Latent Space}

\begin{figure}[htbp]
    \centering
    \includegraphics[width=.8\linewidth]{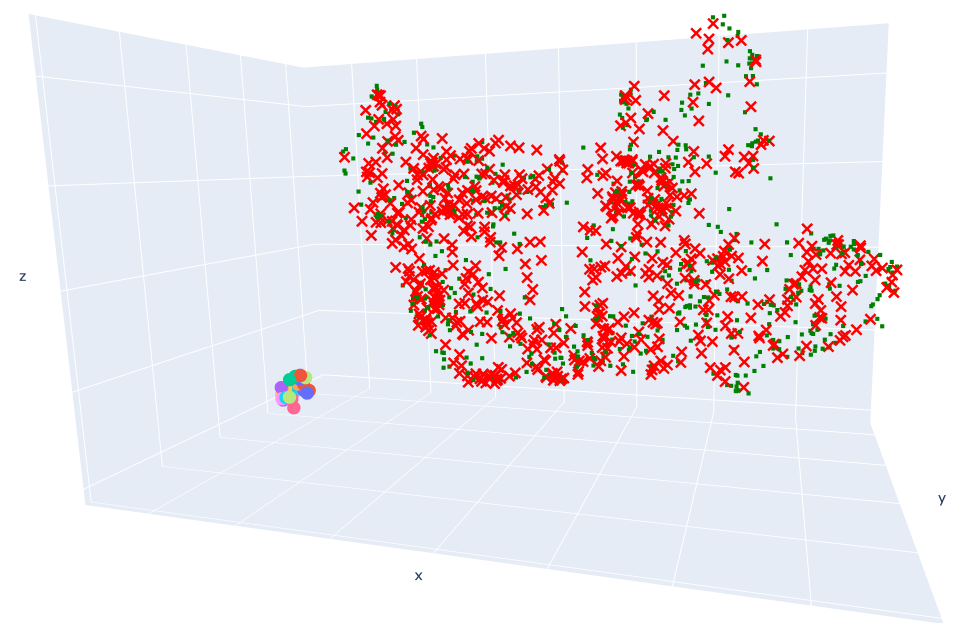}
    \caption{Three-dimensional projection of the textual embeddings of artistic styles and the visual embeddings of correctly classified (green) and misclassified (red) images using UMAP.}
    \label{fig:3D_embedding_graph}
\end{figure}

To investigate the internal organization of the latent space learned by CLIP, we employed the UMAP algorithm \cite{UMAP} to generate a three-dimensional projection of the high-dimensional embeddings. These embeddings included both image representations and textual prompts describing artistic styles (e.g., "an artwork in [style] style"). The resulting visualization (Figure ~\ref{fig:3D_embedding_graph}) reveals a clear separation between textual encodings (the small cluster on the left) and image encodings (the large cluster on the right). For the images, we also distinguish correctly classified samples, shown as green bullets, from misclassified ones, shown as red crosses.

The substantial entanglement of the two image classes suggests a dominance of non-stylistic features in the embeddings. However, the chaotic pattern could also be a consequence of the aggressive dimensionality reduction and may not accurately reflect the semantic structure present in the original high-dimensional space.

To better understand the structure of CLIP’s latent space, we performed a nearest-neighbor analysis, focusing on image pairs where one image was correctly classified and the other was not. These pairs isolate semantically coherent cases with divergent classification outcomes, revealing how subject similarity and stylistic cues interact and where the model struggles to disentangle them.

\begin{figure}[htbp] 
    \centering 
    \begin{tabular}{c} 
    \includegraphics[width=.8\linewidth]{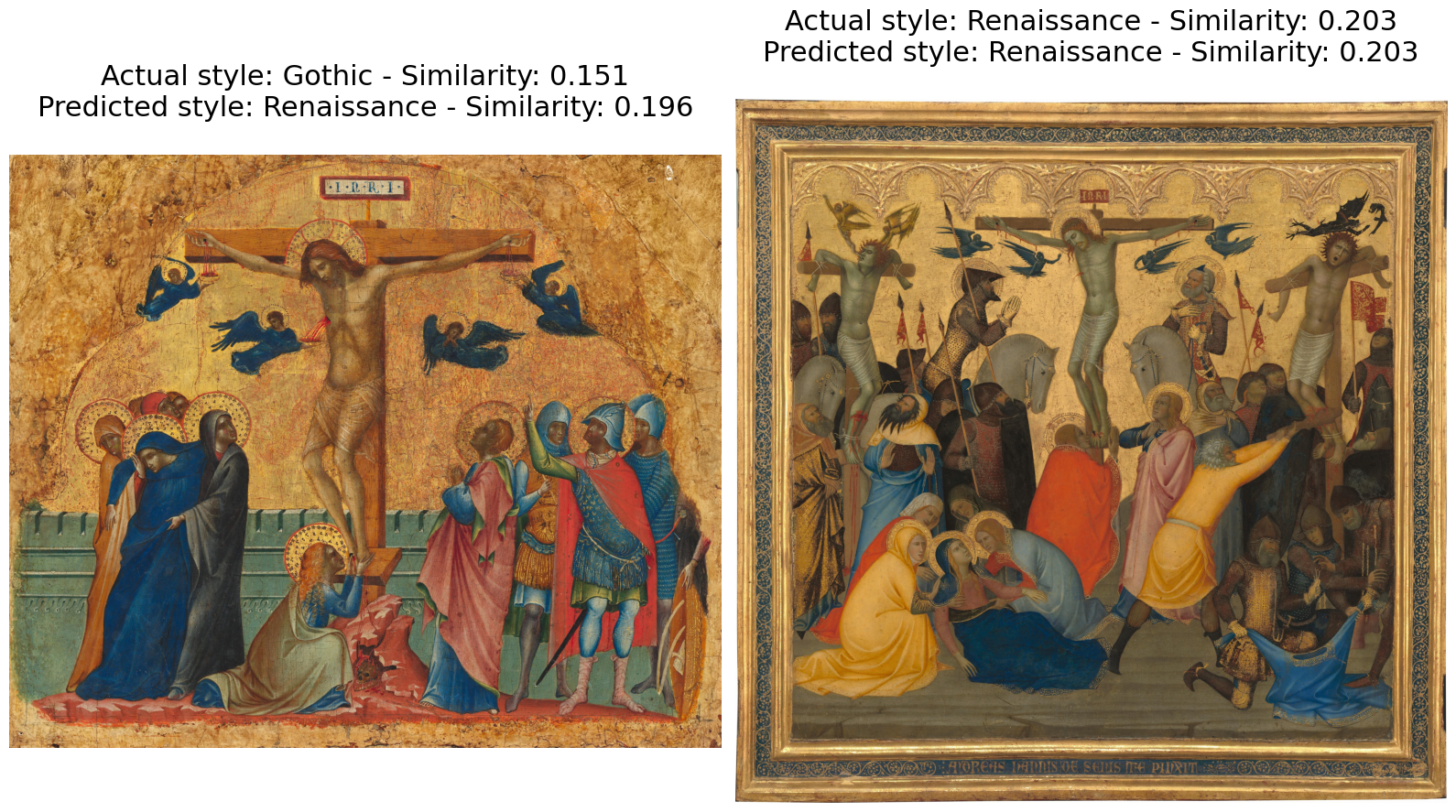} \\
        \includegraphics[width=.8\linewidth]{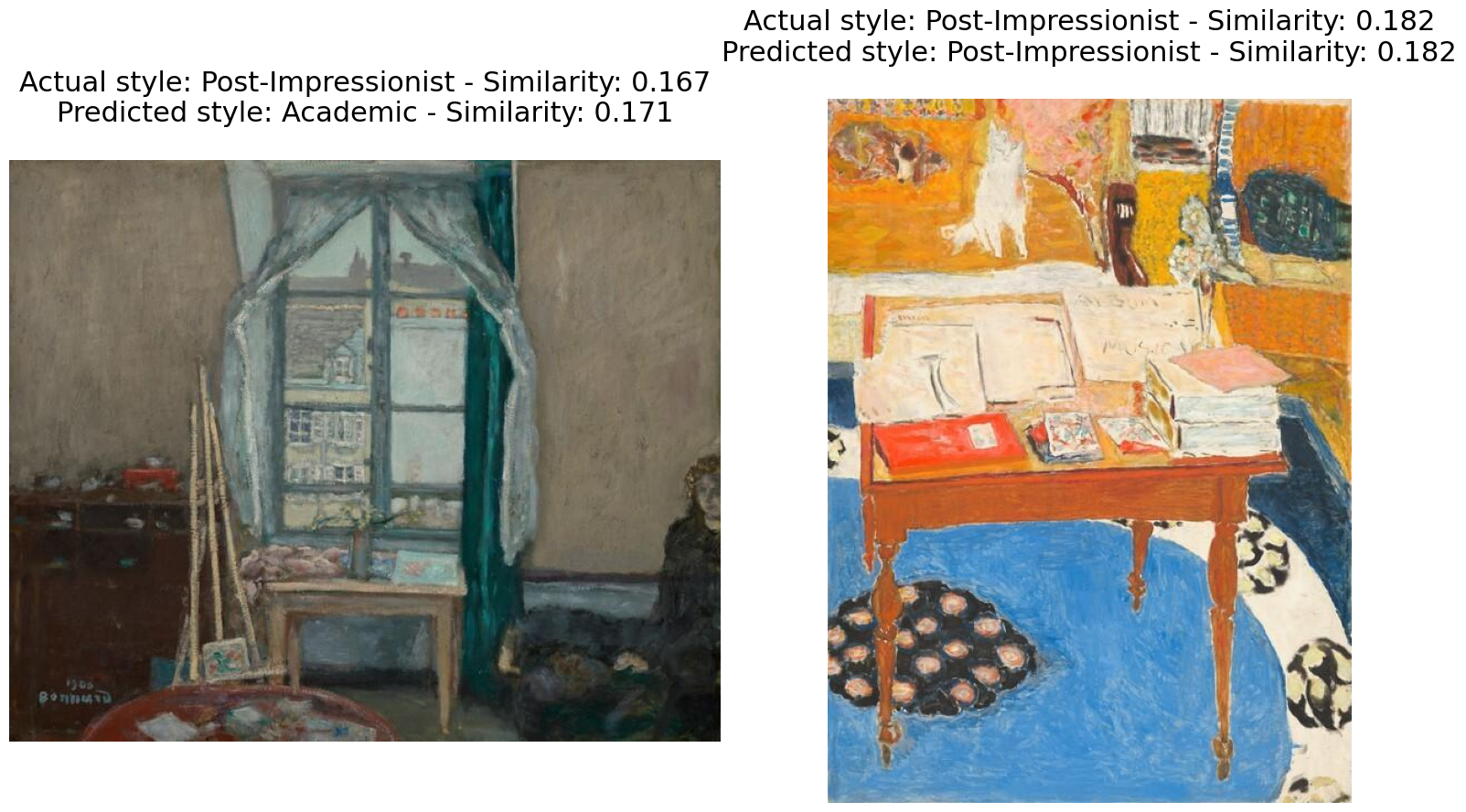} \\ \includegraphics[width=.8\linewidth]{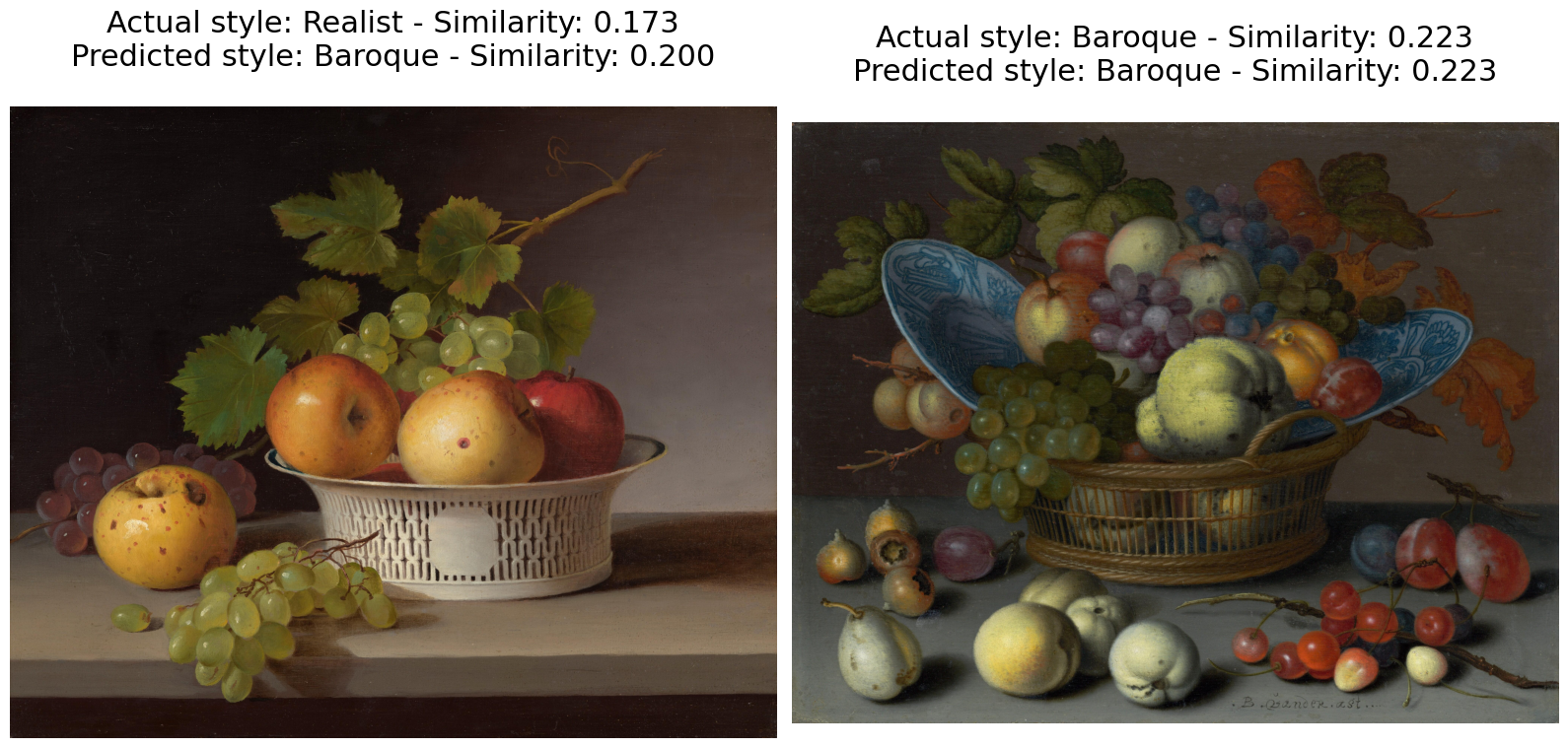} 
    \end{tabular} 
    \caption{Visual comparison between a misclassified image (left) and a correctly classified one (right). For each artwork, the actual and predicted artistic styles are shown, along with their respective similarity scores to the image in CLIP's space.}
    \label{fig:example_style_aligment}
\end{figure}

Representative examples are shown in Figure~\ref{fig:example_style_aligment}, with the misclassified image on the left and its nearest correctly classified neighbor on the right. For each image, cosine similarity scores are reported for both the true and predicted style prompts, in the original latent space and its lower-dimensional projection. This comparison highlights how embedding-space proximity relates to classification outcomes and whether stylistic distinctions are preserved after dimensionality reduction.

These findings highlight a key limitation of CLIP: it prioritizes semantic content  - objects, scenes, and compositions - over stylistic features such as brushwork or color palette. This bias, rooted in training objectives that favor content-based alignment, often leads to misclassifications when artworks share subject matter but differ in style, causing the latent space to conflate stylistically distinct images with similar semantics.


\section{Experiments with AI-generated artworks}

The experiments conducted on AI-generated images follow a structure similar to those applied to human-generated artworks, but with some important caveats. In this case, the prompt used to generate each image serves as the reference summary for computing similarity. Consequently, a low similarity score may indicate a failure on the part of the image generator rather than a shortcoming of CLIP.
\begin{table*}[t]
    \centering
    \begin{tabular}{l|l|l|l|l|l|l|l|l|l}
\textbf{Model} & RN50 & RN101 & RN50x4 & RN50x16 & RN50x64 & ViT-B/32 & ViT-B/16 & ViT-L/14 & ViT-L/14@336px \\ \hline
\multicolumn{1}{c|}{\textbf{Accuracy}} & \multicolumn{1}{c|}{0.866} & \multicolumn{1}{c|}{0.887} & \multicolumn{1}{c|}{0.891} & \multicolumn{1}{c|}{0.893} & \multicolumn{1}{c|}{\textbf{0.896}} & \multicolumn{1}{c|}{0.881} & \multicolumn{1}{c|}{0.880} & \multicolumn{1}{c|}{\textbf{0.896}} & \multicolumn{1}{c}{\textbf{0.896}} 
\end{tabular}
    \caption{Accuracy of different CLIP models in matching generated images with their corresponding summarized prompts in the AI-Pastiche dataset.}
    \label{tb:AI-Pastiche_result_table}
\end{table*}

\begin{table*}[t]
\centering
\begin{tabular}{l|l|l|l|l|l|l|l|l|l}
\textbf{Model} & RN50 & RN101 & RN50x4 & RN50x16 & RN50x64 & ViT-B/32 & ViT-B/16 & ViT-L/14 & ViT-L/14@336px \\ \hline
\textbf{Accuracy} & 0.467 & 0.455 & 0.458 & 0.448 & 0.376 & 0.443 & 0.437 & 0.470 & \textbf{0.487}
\end{tabular}
\caption{Result table for art style recognition on AI-Pastiche}
\label{tb:style_result_table_aipastiche}
\end{table*}

The same ambiguity arises in style classification: in the AI-Pastiche dataset, the ``style" corresponds to the intended style described in the prompt—not necessarily the one successfully rendered in the generated image. As such, any misclassification could be due either to the image generator failing to follow the prompt or to CLIP failing to recognize the intended style. 

To clarify the contribution of these factors, we further compared CLIP’s perception of generated images with human judgments, drawing on user survey data from the AI-Pastiche dataset \cite{AI-pastiche}. Section~\ref{sec:comparison_human} provides details of these experiments.

\subsection{Image-prompt similarity}
\label{sec:prompt similarity}
Here, we are comparing the embedding of the generated image
with the embedding of the relative prompt. 

Embeddings were generated for both images and prompt summaries, and their cosine similarity was calculated. This similarity score was then used to predict the prompt corresponding to each generated image.

Due to the smaller number of textual prompts (72), only accuracy was measured in this case. 
The results are shown in Table \ref{tb:AI-Pastiche_result_table}.
All CLIP models perform well in associating generated images with their corresponding prompt summaries, with accuracy values exceeding 0.86 across the board. The highest performance is achieved by RN50x64, ViT-L/14, and ViT-L/14@336px, all reaching an accuracy of 0.896. The task is sensibly simpler than the image-description alignment of Section \ref{sec:description similarity}, since we only have 73
prompts relative to quite different subjects.
Nevertheless, the results confirm CLIP's ability to capture the visual-semantic correspondence in synthetically generated image-text pairs.

From the perspective of the generators, the high recall indicates that all models perform well in producing images that closely match the subjects described in the prompts. The average cosine similarity between each generated image and its corresponding prompt summary is 0.278, with a standard deviation of 0.344.

Although cosine similarity can be profitably used to associate an image with its prompt, it is not clear if it can be reliably employed as a standalone metric to compare the quality of images obtained from different generators on the same prompt. 
The problem is that this assessment requires a complex evaluation comprising not just the semantic correspondence with the subject but also the stylistic adherence and the technical quality of the generation. This includes evaluating the absence of artifacts, distortions, or visual defects that may not be compatible with the intended artistic style. 


We start addressing stylistic issues in Section~\ref{sec:style_AI-pastiche}, and in Section~\ref{sec:adherence} we will compare the CLIP-evaluation of the adherence between an image and its prompt with a similar evaluation done by humans.

\subsection{Style Recognition}
\label{sec:style_AI-pastiche}
In the second experiment, we used CLIP to evaluate the alignment between generated artworks and their {\em expected} styles, provided among the AI-Pastiche metadata. Similar to the 
case of images from NGAD, we generated a prompt of the 
form "an artwork in [style] style" and computed its cosine similarity with the image embedding. The accuracy results for the different models are shown in Table \ref{tb:style_result_table_ngd}.

\begin{figure}[htbp]
    \centering
    \includegraphics[width=1\linewidth]{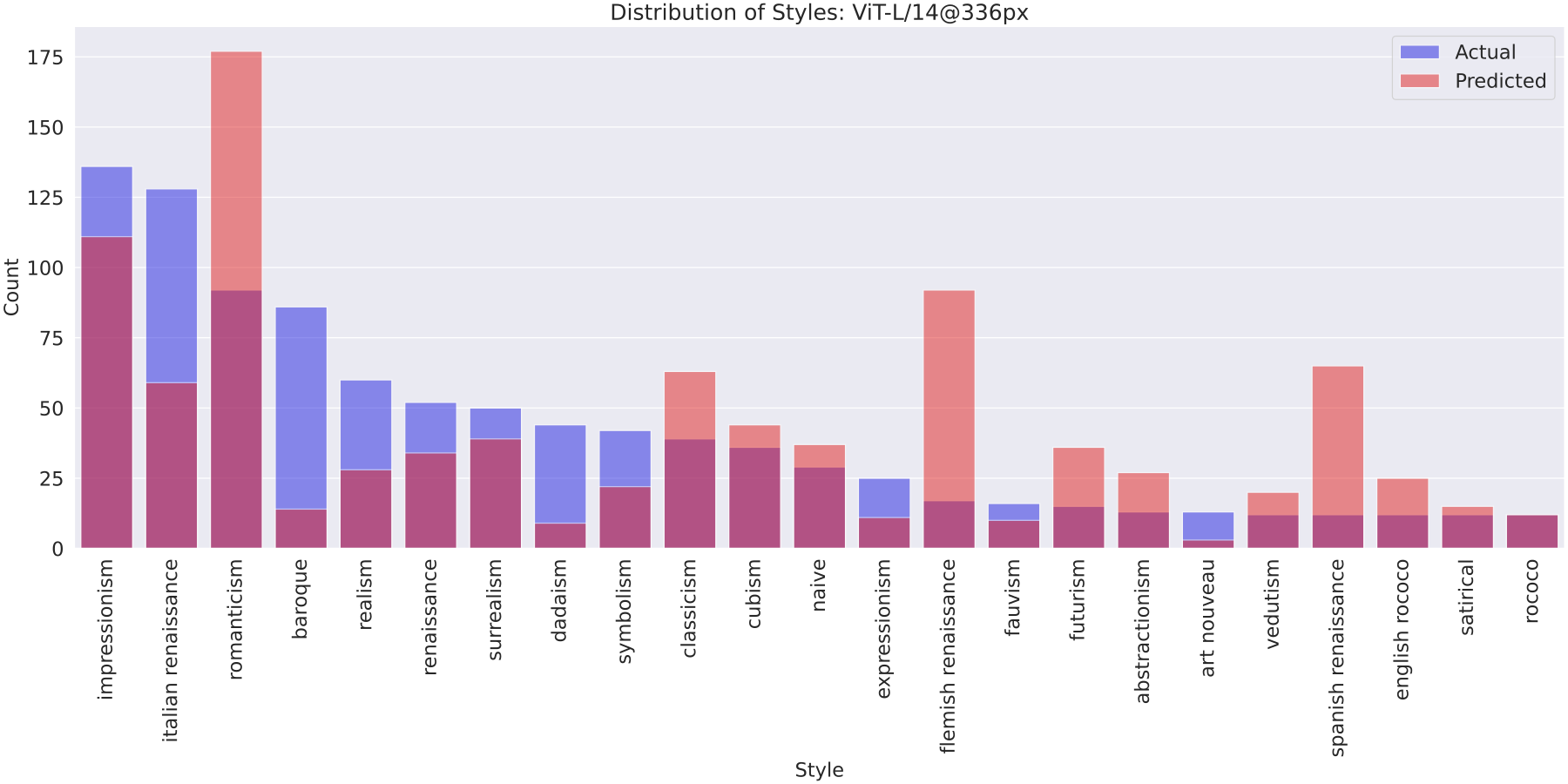}
    \caption{Distribution of actual vs predicted styles in AI-Pastiche}
    \label{fig:style_distribution_aipastiche}
\end{figure}

The distribution of actual and predicted styles in AI-Pastiche (Figure~\ref{fig:style_distribution_aipastiche}) reveals acceptable performance on several of the most frequently prompted styles, such as Renaissance, Impressionism, Surrealism, and Cubism, while styles like Romanticism, Dadaism, and Classicism are more often misclassified.

CLIP achieves a surprisingly high accuracy in predicting the target style of generated images, comparable to that in the NGAD task, despite the fact that, on visual inspection, AI-Pastiche generators often fail to convincingly reproduce the historical styles specified in the prompts. This high agreement may stem from a shared inductive bias, since many generative models integrate CLIP as a scoring function, conditioning mechanism, or similarity guide. To evaluate whether CLIP’s style assessments align with human perception, the adherence and artifact surveys from the AI-Pastiche dataset provide a useful point of comparison.

\section{Comparison with human evaluations}
\label{sec:comparison_human}
This section compares CLIP’s perceptual judgments with human evaluations from two AI-Pastiche surveys: adherence, measuring alignment with prompts, and artifact, assessing visual distortions. 


\subsection{Adherence Analysis}
\label{sec:adherence}
In \cite{AI-pastiche}, an adherence survey asked participants to rate images generated from the same prompt as good, neutral, or bad in terms of prompt adherence. Averaging these ratings produced an adherence score for each image, reflecting perceived stylistic and semantic alignment.

To assess whether CLIP captures similar cues, cosine similarity was computed between the CLIP text embedding of each prompt and the CLIP image embeddings of its generated images. Scores were normalized and scaled to match the human-derived adherence scores.

The correlation between these CLIP scores and human ratings was then calculated as a synthetic measure of alignment, repeated for all available CLIP models. Results appear in the second column of Table~\ref{tb:adhrence_results} (the third column is discussed in Section~\ref{sec:defects}).

\begin{table}[htbp]
\begin{tabular}{p{2.2cm}|p{2.0cm}|p{3.1cm}}
 \textbf{human vs human} & \textbf{CLIP vs human} & \textbf{CLIP + defects vs human}\\\hline
 0.70 & 0.43& 0.50  \\
\end{tabular}
\caption{Correlation between humans' evaluation of the adherence between a generated image and its prompt, and a similar evaluation based on CLIP's (ViT-L/14@336px)}
\label{tb:adhrence_results}
\end{table}

The average correlation between human evaluations of different individuals is around 0.7, while the correlation between
CLIP's adherence assessment with the average human assessment is sensibly lower: 0.43.

\subsection{Perception of Artifacts and Deformations}
\label{sec:defects}
One of the surveys collected in \cite{AI-pastiche} aimed to detect the presence of visible artifacts or deformations in the generated image.
Defects were categorized as Major (clearly visible or frequent errors, such as macroscopic anatomical mistakes), Minor (additional fingers, minor deformations), or None (no apparent mistake).
Results were summarized in a "defect score" associated with each image.

We sought to investigate whether CLIP could identify and detect these kinds of mistakes. Our initial investigation aimed to determine if we could approximate the defect score through linear regression, starting from the CLIP embedding of the AI-Pastiche images. 
The result was negative: we obtained a coefficient of determination $R^2$ close to 0. 

As additional evidence that CLIP embeddings do not account for defects and artifacts in input images, we tested whether CLIP’s evaluation of prompt adherence could be improved by incorporating a linear combination with the human-evaluated "defect score". This turns out to be the case: a suitable linear combination achieves a similarity of nearly 0.5 with the average human adherence evaluation (see the third column in Table~\ref{tb:adhrence_results}).

\begin{figure}[htbp]
 \begin{tabular}{ccc}
    \includegraphics[height=.29\columnwidth]{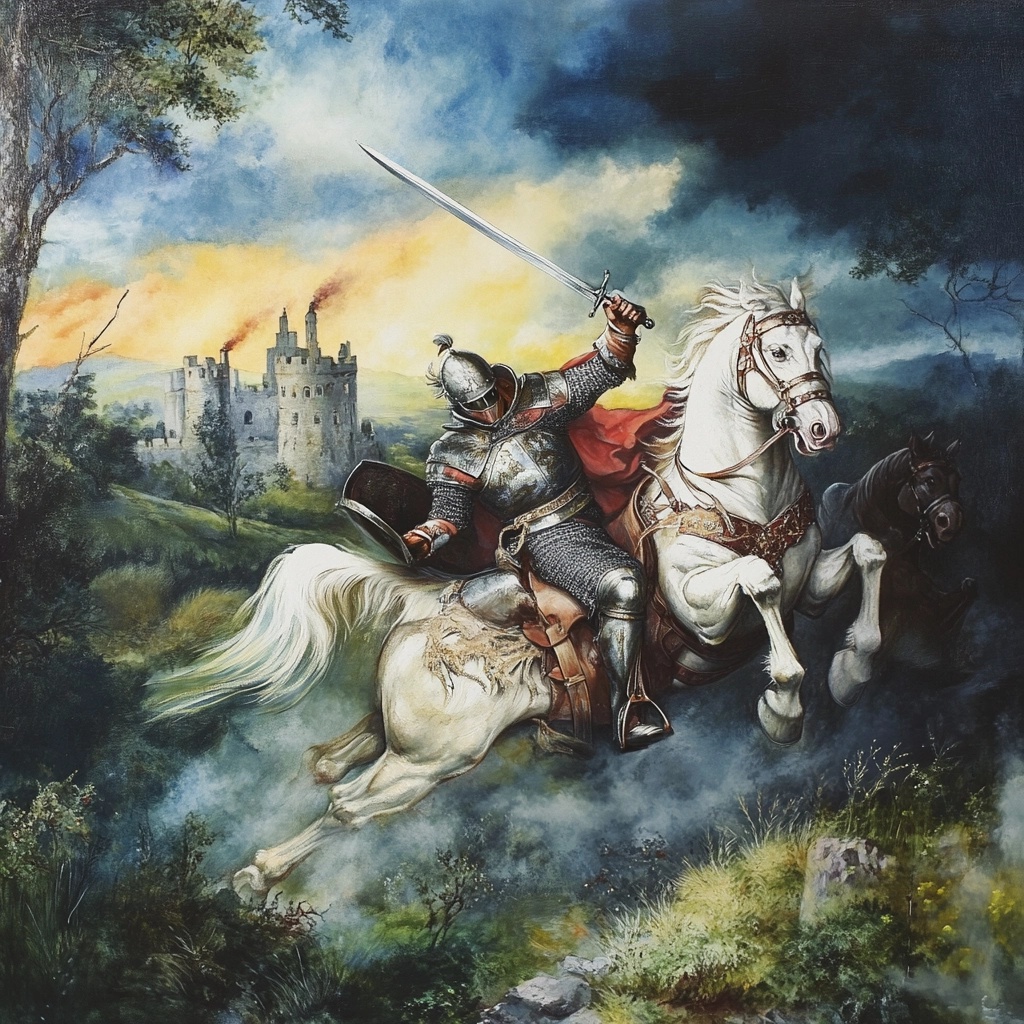} &
     \includegraphics[height=.29\columnwidth]{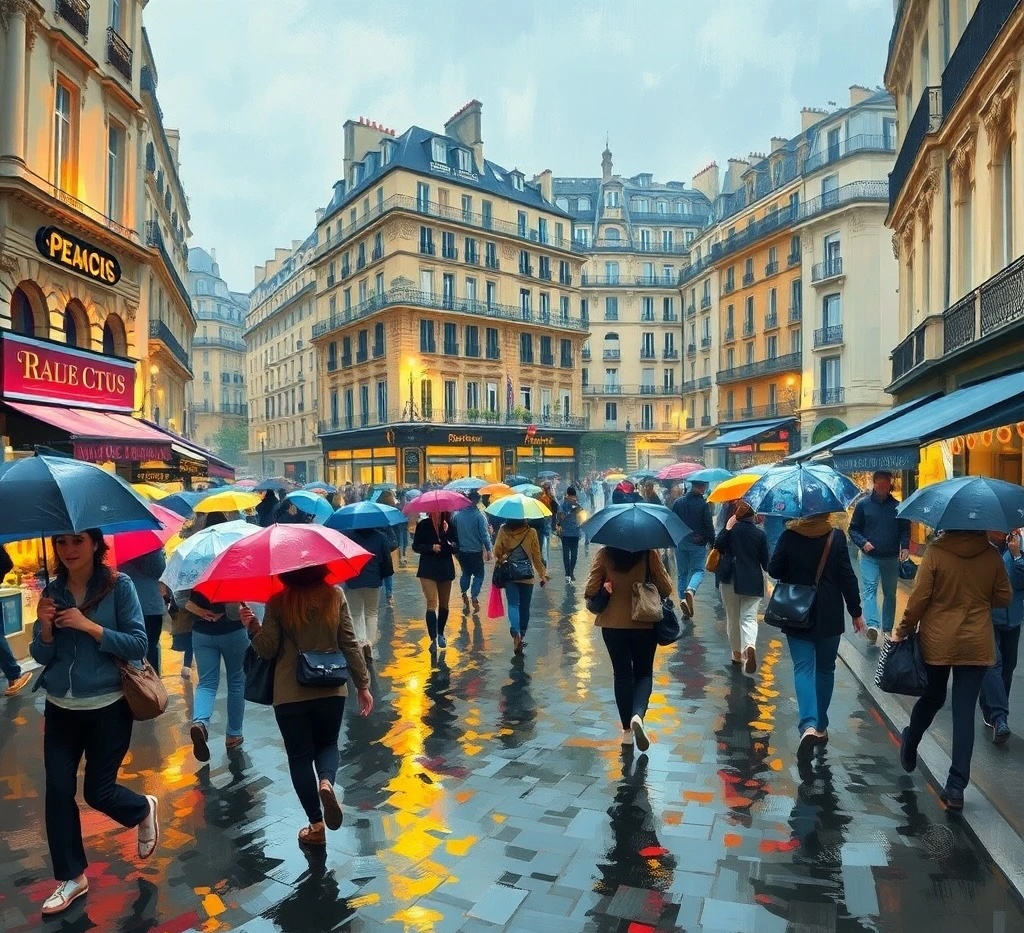} &     \includegraphics[height=.29\columnwidth]{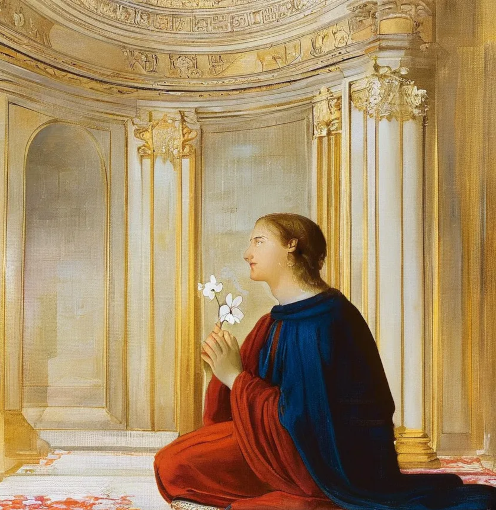}
     \\
     (a) Midjourney & (b) Auto-Aesthetics & (c) Omnigen
 \end{tabular}
 \caption{Examples of images in the AI AI-Pastiche dataset not aligning well with their prompts, for content or style.}
 \label{fig:examples_AIpastiche}
\end{figure}

The remaining mistakes in CLIP's perception seem to be mainly related to counting problems or stylistic issues.
Some illustrative examples are shown in Figure~\ref{fig:examples_AIpastiche}.

Figure (a) was intended to depict two knights fighting on horseback; humans penalized the absence of one knight, which CLIP overlooked. Image (b), from Auto-Aesthetics V1, was meant to show a rainy Parisian street in an impressionist style; humans criticized its stylistic inaccuracy. For image (c), the prompt specified a kneeling figure in rich robes offering a white flower to another figure in blue; CLIP similarity did not register the absence of one figure.

\section{Conclusion}
Our investigation into CLIP’s perception of artworks, conducted across both human-created and AI-generated images, reveals a model with remarkable breadth, but still far from capturing the richness of human aesthetics and contextual understanding. While CLIP is adept at grounding images in broad semantic categories and descriptive summaries, it often falters when asked to navigate the more subjective terrain of artistic nuance: style, composition, and technical accuracy.

Both the vision system and the cosine similarity measure have limitations. The vision system often fails to detect the presence and location of artifacts or defects. At the same time, extracting rich stylistic information through textual embeddings and simple cosine similarity is a challenge. This underuse of the visual representation is particularly problematic in art, where images convey signals - composition, narrative, emotional tone - that demand a deep semantic connection to their textual description.

Looking ahead, future vision–language systems must go beyond mere alignment. What is needed is a deeper model of perception, one that can reason about images in terms of objects, styles, historical context, artistic intent, and visual storytelling. Achieving this may require richer supervision, incorporating not only captioned data but also art-historical metadata, expert narratives, and multimodal dialogues.

As AI takes on a greater role in creating, curating, and critiquing visual culture, the question becomes not just what models see, but how they see, and whose eyes they are borrowing. Current vision mechanisms are useful but partial lenses, limited in capturing the depth and nuance of human artistic perception. This study exposes such constraints in CLIP, highlighting the need to examine whether more advanced multimodal systems, including LLM-based architectures, can overcome these limitations or simply reproduce them in subtler forms.


\begin{thebibliography}{10}

\bibitem{CLIP_limitations}
Reza Abbasi, Ali Nazari, Aminreza Sefid, Mohammadali Banayeeanzade, Mohammad~Hossein Rohban, and Mahdieh~Soleymani Baghshah.
\newblock Analyzing clip's performance limitations in multi-object scenarios: A controlled high-resolution study.
\newblock {\em arXiv preprint arXiv:2502.19828}, 2025.

\bibitem{does_clip_arxiv}
Andrea Asperti, Leonardo Dess{\`{\i}}, Maria~Chiara Tonetti, and Nico Wu.
\newblock Does {CLIP} perceive art the same way we do?
\newblock {\em CoRR}, abs/2505.05229, 2025.

\bibitem{AI-pastiche}
Andrea Asperti, Franky George, Tiberio Marras, Razvan~Ciprian Stricescu, and Fabio Zanotti.
\newblock A critical assessment of modern generative models’ ability to replicate artistic styles.
\newblock {\em Big Data and Cognitive Computing}, 9(9), 2025.

\bibitem{CLIP_GAP}
Andrea Asperti, Leonardo Naldi, and Salvatore Fiorilla.
\newblock An investigation of the domain gap in clip-based person re-identification.
\newblock {\em Sensors}, 25(2), 2025.

\bibitem{Retrieval_CLIP}
Alberto Baldrati, Marco Bertini, Tiberio Uricchio, and Alberto~Del Bimbo.
\newblock Composed image retrieval using contrastive learning and task-oriented clip-based features.
\newblock {\em {ACM} Trans. Multim. Comput. Commun. Appl.}, 20(3):62:1--62:24, 2024.

\bibitem{inverting_CNN}
Alexey Dosovitskiy and Thomas Brox.
\newblock Inverting visual representations with convolutional networks.
\newblock In {\em 2016 {IEEE} Conference on Computer Vision and Pattern Recognition, {CVPR} 2016, Las Vegas, NV, USA, June 27-30, 2016}, pages 4829--4837. {IEEE} Computer Society, 2016.

\bibitem{stable_diffusion_3_5_large_t2i}
Patrick Esser, Sumith Kulal, Andreas Blattmann, Rahim Entezari, Jonas M{\"{u}}ller, Harry Saini, Yam Levi, Dominik Lorenz, Axel Sauer, Frederic Boesel, Dustin Podell, Tim Dockhorn, Zion English, Kyle Lacey, Alex Goodwin, Yannik Marek, and Robin Rombach.
\newblock Scaling rectified flow transformers for high-resolution image synthesis.
\newblock {\em CoRR}, abs/2403.03206, 2024.

\bibitem{CLIPDraw}
Kevin Frans, Lisa~B. Soros, and Olaf Witkowski.
\newblock Clipdraw: Exploring text-to-drawing synthesis through language-image encoders.
\newblock In Sanmi Koyejo, S.~Mohamed, A.~Agarwal, Danielle Belgrave, K.~Cho, and A.~Oh, editors, {\em Advances in Neural Information Processing Systems 35: Annual Conference on Neural Information Processing Systems 2022, NeurIPS 2022, New Orleans, LA, USA, November 28 - December 9, 2022}, 2022.

\bibitem{CLIP-adapter}
Peng Gao, Shijie Geng, Renrui Zhang, Teli Ma, Rongyao Fang, Yongfeng Zhang, Hongsheng Li, and Yu~Qiao.
\newblock Clip-adapter: Better vision-language models with feature adapters.
\newblock {\em Int. J. Comput. Vis.}, 132(2):581--595, 2024.

\bibitem{Distilling_CLIP}
Zeyi Huang, Andy Zhou, Zijian Lin, Mu~Cai, Haohan Wang, and Yong~Jae Lee.
\newblock A sentence speaks a thousand images: Domain generalization through distilling {CLIP} with language guidance.
\newblock In {\em {IEEE/CVF} International Conference on Computer Vision, {ICCV} 2023, Paris, France, October 1-6, 2023}, pages 11651--11661, 2023.

\bibitem{inverting_CLIP}
Hamid Kazemi, Atoosa~Malemir Chegini, Jonas Geiping, Soheil Feizi, and Tom Goldstein.
\newblock What do we learn from inverting {CLIP} models?
\newblock {\em CoRR}, abs/2403.02580, 2024.

\bibitem{CLIPSAM}
Shengze Li, Jianjian Cao, Peng Ye, Yuhan Ding, Chongjun Tu, and Tao Chen.
\newblock Clipsam: {CLIP} and {SAM} collaboration for zero-shot anomaly segmentation.
\newblock {\em Neurocomputing}, 618:129122, 2025.

\bibitem{CLIP-ReID}
Siyuan Li, Li~Sun, and Qingli Li.
\newblock Clip-reid: Exploiting vision-language model for image re-identification without concrete text labels.
\newblock In {\em Thirty-Seventh {AAAI} Conference on Artificial Intelligence, {AAAI} 2023, Thirty-Fifth Conference on Innovative Applications of Artificial Intelligence, {IAAI} 2023, Thirteenth Symposium on Educational Advances in Artificial Intelligence, {EAAI} 2023, Washington, DC, USA, February 7-14, 2023}, pages 1405--1413, 2023.

\bibitem{Semantic_Diffusion_Guidance}
Xihui Liu, Dong~Huk Park, Samaneh Azadi, Gong Zhang, Arman Chopikyan, Yuxiao Hu, Humphrey Shi, Anna Rohrbach, and Trevor Darrell.
\newblock More control for free! image synthesis with semantic diffusion guidance.
\newblock In {\em {IEEE/CVF} Winter Conference on Applications of Computer Vision, {WACV} 2023, Waikoloa, HI, USA, January 2-7, 2023}, pages 289--299. {IEEE}, 2023.

\bibitem{CLIP_ECAI2024}
Zhihang Liu, Qiang Huang, and Yingying Zhu.
\newblock Clip-based cross-level semantic interaction and recombination network for composed image retrieval.
\newblock In {\em {ECAI} 2024 - 27th European Conference on Artificial Intelligence, 19-24 October 2024, Santiago de Compostela, Spain - Including 13th Conference on Prestigious Applications of Intelligent Systems {(PAIS} 2024)}, pages 704--711, 2024.

\bibitem{DBLP:conf/colins/LytvynPRKG24}
Vasyl Lytvyn, Roman Peleshchak, Ihor Rishnyak, Bohdan Kopach, and Yuriy Gal.
\newblock Detection of similarity between images based on contrastive language-image pre-training neural network.
\newblock In Vasyl Lytvyn, Agnieszka Kowalska{-}Styczen, and Victoria Vysotska, editors, {\em Proceedings of the 8th International Conference on Computational Linguistics and Intelligent Systems. Volume {I:} Machine Learning Workshop, Lviv, Ukraine, April 12-13, 2024}, volume 3664 of {\em {CEUR} Workshop Proceedings}, pages 94--104. CEUR-WS.org, 2024.

\bibitem{Vevaldi15}
Aravindh Mahendran and Andrea Vedaldi.
\newblock Understanding deep image representations by inverting them.
\newblock In {\em {IEEE} Conference on Computer Vision and Pattern Recognition, {CVPR} 2015, Boston, MA, USA, June 7-12, 2015}, pages 5188--5196. {IEEE} Computer Society, 2015.

\bibitem{UMAP}
Leland McInnes and John Healy.
\newblock {UMAP:} uniform manifold approximation and projection for dimension reduction.
\newblock {\em CoRR}, abs/1802.03426, 2018.

\bibitem{GLIDE}
Alexander~Quinn Nichol, Prafulla Dhariwal, Aditya Ramesh, Pranav Shyam, Pamela Mishkin, Bob McGrew, Ilya Sutskever, and Mark Chen.
\newblock {GLIDE:} towards photorealistic image generation and editing with text-guided diffusion models.
\newblock In {\em International Conference on Machine Learning, {ICML} 2022, 17-23 July 2022, Baltimore, Maryland, {USA}}, pages 16784--16804, 2022.

\bibitem{nga_real_dataset}
National~Gallery of~Art.
\newblock National gallery of art open data program, January 2024.
\newblock Accessed: 2024-01-29.

\bibitem{CLIP_semantic_knowledge}
Xuran Pan, Tianzhu Ye, Dongchen Han, Shiji Song, and Gao Huang.
\newblock Contrastive language-image pre-training with knowledge graphs.
\newblock In Sanmi Koyejo, S.~Mohamed, A.~Agarwal, Danielle Belgrave, K.~Cho, and A.~Oh, editors, {\em Advances in Neural Information Processing Systems 35: Annual Conference on Neural Information Processing Systems 2022, NeurIPS 2022, New Orleans, LA, USA, November 28 - December 9, 2022}, 2022.

\bibitem{SgVA-CLIP}
Fang Peng, Xiaoshan Yang, Linhui Xiao, Yaowei Wang, and Changsheng Xu.
\newblock Sgva-clip: Semantic-guided visual adapting of vision-language models for few-shot image classification.
\newblock {\em {IEEE} Trans. Multim.}, 26:3469--3480, 2024.

\bibitem{CLIP}
Alec Radford, Jong~Wook Kim, Chris Hallacy, Aditya Ramesh, Gabriel Goh, Sandhini Agarwal, Girish Sastry, Amanda Askell, Pamela Mishkin, Jack Clark, Gretchen Krueger, and Ilya Sutskever.
\newblock Learning transferable visual models from natural language supervision.
\newblock In Marina Meila and Tong Zhang, editors, {\em Proceedings of the 38th International Conference on Machine Learning, {ICML} 2021, 18-24 July 2021, Virtual Event}, volume 139 of {\em Proceedings of Machine Learning Research}, pages 8748--8763, \hspace{0pt}, 2021. {PMLR}.
\newblock Accessed: 2025-02-13.

\bibitem{DALLE2}
Aditya Ramesh, Prafulla Dhariwal, Alex Nichol, Casey Chu, and Mark Chen.
\newblock Hierarchical text-conditional image generation with clip latents.
\newblock {\em arXiv e-prints}, pages arXiv--2204, 2022.

\bibitem{stable-diffusion}
Robin Rombach, Andreas Blattmann, Dominik Lorenz, Patrick Esser, and Bj{\"o}rn Ommer.
\newblock High-resolution image synthesis with latent diffusion models.
\newblock In {\em Proceedings of the IEEE/CVF Conference on Computer Vision and Pattern Recognition}, pages 10684--10695, 2022.

\bibitem{CLIP_adversarial}
Vincenzo~De Rosa, Fabrizio Guillaro, Giovanni Poggi, Davide Cozzolino, and Luisa Verdoliva.
\newblock Exploring the adversarial robustness of {CLIP} for ai-generated image detection.
\newblock In {\em {IEEE} International Workshop on Information Forensics and Security, {WIFS} 2024, Rome, Italy, December 2-5, 2024}, pages 1--6. {IEEE}, 2024.

\bibitem{context_aware_pretraining}
Karsten Roth, Zeynep Akata, Dima Damen, Ivana Balazevic, and Olivier~J. H{\'{e}}naff.
\newblock Context-aware multimodal pretraining.
\newblock {\em CoRR}, abs/2411.15099, 2024.

\bibitem{CLIP_prior_guidance}
Yogesh Surapaneni and Chakravarthy Bhagvati.
\newblock Scene text image super-resolution with {CLIP} prior guidance.
\newblock In {\em Pattern Recognition - 27th International Conference, {ICPR} 2024, Kolkata, India, December 1-5, 2024, Proceedings, Part {XXXII}}, pages 17--32, 2024.

\bibitem{CLIP_closer_look}
Ashkan Taghipour, Morteza Ghahremani, Mohammed Bennamoun, Aref~Miri Rekavandi, Zinuo Li, Hamid Laga, and Farid Boussa{\"{\i}}d.
\newblock Faster image2video generation: {A} closer look at {CLIP} image embedding's impact on spatio-temporal cross-attentions.
\newblock {\em CoRR}, abs/2407.19205, 2024.

\bibitem{CLIP_robustness}
Weijie Tu, Weijian Deng, and Tom Gedeon.
\newblock Toward a holistic evaluation of robustness in {CLIP} models.
\newblock {\em CoRR}, abs/2410.01534, 2024.

\bibitem{CLIPMulti}
Peng Wang, Dagang Li, Xuesi Hu, Yongmei Wang, and Youhua Zhang.
\newblock Clipmulti: Explore the performance of multimodal enhanced {CLIP} for zero-shot text classification.
\newblock {\em Comput. Speech Lang.}, 90:101748, 2025.

\bibitem{CAMIR}
Fan Yang, Nor~Azman Ismail, Pang~Yee Yong, and Alhuseen~Omar Alsayed.
\newblock {CAMIR:} fine-tuning {CLIP} and multi-head cross-attention mechanism for multimodal image retrieval with sketch and text features.
\newblock {\em Int. J. Multim. Inf. Retr.}, 14(1):2, 2025.

\bibitem{CLIP-zero-shot}
Hairui Yang, Ning Wang, Haojie Li, Lei Wang, and Zhihui Wang.
\newblock Application of {CLIP} for efficient zero-shot learning.
\newblock {\em Multim. Syst.}, 30(4):219, 2024.

\bibitem{dreaming_to_distill}
Hongxu Yin, Pavlo Molchanov, Jos{\'{e}}~M. {\'{A}}lvarez, Zhizhong Li, Arun Mallya, Derek Hoiem, Niraj~K. Jha, and Jan Kautz.
\newblock Dreaming to distill: Data-free knowledge transfer via deepinversion.
\newblock In {\em 2020 {IEEE/CVF} Conference on Computer Vision and Pattern Recognition, {CVPR} 2020, Seattle, WA, USA, June 13-19, 2020}, pages 8712--8721. Computer Vision Foundation / {IEEE}, 2020.

\bibitem{TIP-adapter}
Tao Yu, Zhihe Lu, Xin Jin, Zhibo Chen, and Xinchao Wang.
\newblock Task residual for tuning vision-language models.
\newblock In {\em {IEEE/CVF} Conference on Computer Vision and Pattern Recognition, {CVPR} 2023, Vancouver, BC, Canada, June 17-24, 2023}, pages 10899--10909. {IEEE}, 2023.

\bibitem{CoCoOp}
Kaiyang Zhou, Jingkang Yang, Chen~Change Loy, and Ziwei Liu.
\newblock { Conditional Prompt Learning for Vision-Language Models }.
\newblock In {\em 2022 IEEE/CVF Conference on Computer Vision and Pattern Recognition (CVPR)}, pages 16795--16804, Los Alamitos, CA, USA, June 2022. IEEE Computer Society.

\bibitem{CoOp}
Kaiyang Zhou, Jingkang Yang, Chen~Change Loy, and Ziwei~Liu Liu.
\newblock Learning to prompt for vision-language models.
\newblock {\em International Journal of Computer Vision}, 130:2337–2348, 2022.

\bibitem{CLIP-MUSED}
Qiongyi Zhou, Changde Du, Shengpei Wang, and Huiguang He.
\newblock {CLIP-MUSED:} clip-guided multi-subject visual neural information semantic decoding.
\newblock In {\em The Twelfth International Conference on Learning Representations, {ICLR} 2024, Vienna, Austria, May 7-11, 2024}, 2024.

\bibitem{APE}
Xiangyang Zhu, Renrui Zhang, Bowei He, Aojun Zhou, Dong Wang, Bin Zhao, and Peng Gao.
\newblock Not all features matter: Enhancing few-shot {CLIP} with adaptive prior refinement.
\newblock In {\em {IEEE/CVF} International Conference on Computer Vision, {ICCV} 2023, Paris, France, October 1-6, 2023}, pages 2605--2615. {IEEE}, 2023.

\bibitem{CLIP_subspace}
Xingyu Zhu, Beier Zhu, Yi~Tan, Shuo Wang, Yanbin Hao, and Hanwang Zhang.
\newblock Selective vision-language subspace projection for few-shot {CLIP}.
\newblock In Jianfei Cai, Mohan~S. Kankanhalli, Balakrishnan Prabhakaran, Susanne Boll, Ramanathan Subramanian, Liang Zheng, Vivek~K. Singh, Pablo C{\'{e}}sar, Lexing Xie, and Dong Xu, editors, {\em Proceedings of the 32nd {ACM} International Conference on Multimedia, {MM} 2024, Melbourne, VIC, Australia, 28 October 2024 - 1 November 2024}, pages 3848--3857. {ACM}, 2024.

\end{thebibliography}

\end{document}